\documentclass[lettersize,journal]{IEEEtran}

\IEEEoverridecommandlockouts
\usepackage{cite}
\usepackage{amsmath,amssymb,amsfonts}
\usepackage{graphicx}
\usepackage{textcomp}
\usepackage{xcolor}
\usepackage{multirow}
\usepackage{algorithm}
\usepackage{color}
\usepackage{threeparttable}
\usepackage{booktabs} 
\usepackage{hyperref}
\usepackage{graphicx}
\usepackage{subfigure}
\usepackage{mathtools}
\usepackage{xcolor}
\usepackage{colortbl}
\usepackage{bm}
\usepackage{booktabs}
\usepackage{makecell}
\IEEEoverridecommandlockouts
\usepackage[shortcuts,acronym]{glossaries}
\usepackage{multicol} 
\usepackage{cancel}
\usepackage{algpseudocode}
\usepackage[capitalize]{cleveref}

\newacronym{ofdm}{OFDM}{Orthogonal Frequency-Division Multiplexing}
\newacronym{ngp}{Instant-NGP}{Instant Neural Graphics Primitives}
\newacronym{nerf}{NeRF}{Neural Radiance Field}
\newacronym{cv}{CV}{Computer Vision}
\newacronym{psnr}{PSNR}{peak signal-to-noise ratio}
\newacronym{ssim}{SSIM}{structural similarity index measure}
\newacronym{spp}{SPP}{switched Poisson process}
\newacronym{ugv}{UGV}{unmanned ground vehicle}
\newacronym{uav}{UAV}{unmanned aerial vehicle}
\newacronym{bs}{BS}{base station}
\newacronym{lpips}{LPIPS}{learned perceptual image patch similarity}
\newacronym{mse}{MSE}{mean squared error}
\newacronym{cuda}{CUDA}{Compute Unified Device Architecture}
\newacronym{ge}{G-E}{Gilbert-Elliot}
\newacronym{mmpp}{MMPP}{Markov-modulated Poisson Process}
\newacronym{qos}{QoS}{Quality of Service}
\newacronym{aoi}{AoI}{Age of Information}
\newacronym{drl}{DRL}{Deep Reinforcement Learning}
\newacronym{voi}{VoI}{Value of Information}
\newacronym{ros}{ROS}{Robot Operating System}
\newacronym{ppo}{PPO}{Proximal Policy Optimization}
\newacronym{bleu}{BLEU}{Bilingual Evaluation Understudy}
\newacronym{mat}{MAT}{Maximum AoI Threshold}
\newacronym{mlp}{MLP}{multilayer perceptron}
\newacronym{aaoi}{aAoI}{average Age of Information}
\newacronym{3d}{3D}{Three-dimensional}
\newacronym{2d}{2D}{two-dimensional}
\newacronym{vr}{VR}{Virtual Reality}
\newacronym{ar}{AR}{Augmented Reality}
\newacronym{mr}{MR}{Mixed Reality}
\newacronym{3dgs}{3DGS}{3D Gaussian Splatting}
\newacronym{rl}{RL}{Reinforcement Learning}
\newacronym{kpi}{KPI}{Key Performance Indicator}
\newacronym{ai}{AI}{Artificial Intelligence}
\newacronym{gs}{GS}{Gaussian Splatting}
\newacronym{slam}{SLAM}{Simultaneous Localization and Mapping}
\newacronym{rt}{RT}{real-time}
\newacronym{qoe}{QoE}{Quality of Experience}
\newacronym{tosc}{TOSC}{Task-Oriented Semantic Communication}
\newacronym{jscc}{JSCC}{Joint Source-Channel Coding}

\def\BibTeX{{\rm B\kern-.05em{\sc i\kern-.025em b}\kern-.08em
    T\kern-.1667em\lower.7ex\hbox{E}\kern-.125emX}}
\begin{document}

\title{Task-Oriented Communications for 3D Scene Representation: Balancing Timeliness and Fidelity \\

}

\author{
  \IEEEauthorblockN{
                      Xiangmin Xu\textsuperscript{1}, \textit{Student Member}, \textit{IEEE},
                      Zhen Meng\textsuperscript{1},
                      Kan Chen\textsuperscript{1},
                      \textit{Student Member}, \textit{IEEE},
                      Jiaming Yang\textsuperscript{1},
                      \textit{Student Member}, \textit{IEEE},
                      Emma Li\textsuperscript{1}, 
                      Philip G. Zhao\textsuperscript{2},
                      \textit{Senior Member}, \textit{IEEE}, and
                      David Flynn\textsuperscript{3},
                      \textit{Senior Member}, \textit{IEEE},}
                      
\thanks{$^{1}$Xiangmin Xu, Zhen Meng, Kan Chen, Jiaming Yang, and Emma Li are with the School of Computing Science, University of Glasgow, G12 8RZ, Glasgow, UK.
        {\tt\small \{x.xu.1, k.chen.1, j.yang.6\}@research.gla.ac.uk, \{zhen.meng, liying.li\}@glasgow.ac.uk.}}%
\thanks{$^{2}$Philip Zhao is with the Department of Computer Science, University of Manchester, M13 9PL, Manchester, UK.
        {\tt\small philip.zhao@manchester.ac.uk.}}
\thanks{$^{3}$David Flynn is with the School of Engineering, University of Glasgow, G12 8QQ, Glasgow, UK. {\tt\small david.flynn@glasgow.ac.uk.}}

}

\maketitle
 
\begin{abstract}
Real-time Three-dimensional (3D) scene representation is a foundational element that supports a broad spectrum of cutting-edge applications, including digital manufacturing, Virtual, Augmented, and Mixed Reality (VR/AR/MR), and the emerging metaverse. Despite advancements in real-time communication and computing, achieving a balance between timeliness and fidelity in 3D scene representation remains a challenge. This work investigates a wireless network where multiple homogeneous mobile robots, equipped with cameras, capture an environment and transmit images to an edge server over channels for 3D representation. We propose a contextual-bandit Proximal Policy Optimization (PPO) framework incorporating both Age of Information (AoI) and semantic information to optimize image selection for representation, balancing data freshness and representation quality. Two policies—the $\omega$-threshold and $\omega$-wait policies—together with two benchmark methods are evaluated, timeliness embedding and weighted sum, on standard datasets and baseline 3D scene representation models.
Experimental results demonstrate improved representation fidelity while maintaining low latency, offering insight into the model's decision-making process. This work advances real-time 3D scene representation by optimizing the trade-off between timeliness and fidelity in dynamic environments.

\end{abstract}


\section{Introduction}
\gls{3d} scene representations encompass the processes of capturing, interpreting, and reconstructing \gls{3d} objects from \gls{2d} images or sensor data~\cite{tewari2022advances}. These representations are foundational across diverse applications, including digital manufacturing\cite{industrialnerf}, autonomous driving\cite{feifeidriving}, \gls{vr}, \gls{ar}, \gls{mr}\cite{aoipetar}, and the metaverse\cite{zhenjsac}. For instance, immersive metaverse worlds rely on the creation of detailed \gls{3d} virtual objects, such as buildings and vegetation, from sparse \gls{2d} images~\cite{meng2023taskoriented}.
\begin{figure}
    \centering
    \includegraphics[width=\linewidth]{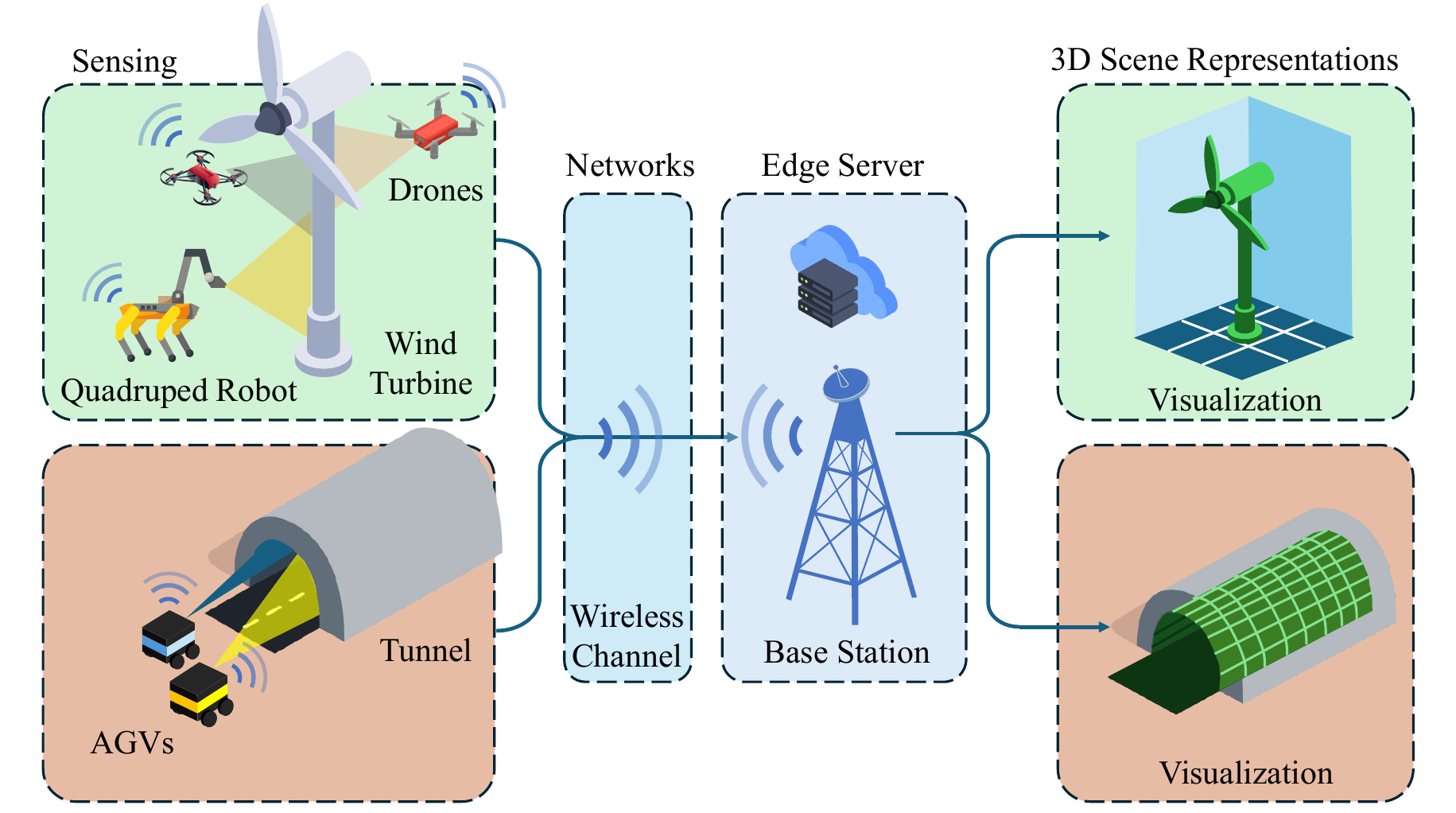}
    \caption{System model of edge-assisted 3D scene reconstruction, where heterogeneous sensors (e.g., drones, quadruped robots, and AGVs) capture data from industrial environments such as wind turbines and tunnels. The sensed data are transmitted via wireless channels to the edge server through a base station, and then processed for real-time 3D scene visualization.}
    \label{fig: poster}
\end{figure}
Traditional explicit representation methods such as point clouds~\cite{qi2017pointnet}, voxel grids~\cite{maturana2015voxnet}, and mesh reconstruction~\cite{kazhdan2006poisson} have long been used to generate \gls{3d} models. However, these approaches present limitations, particularly in real-time contexts, due to challenges like computational overhead, storage demands, and difficulty in handling sparse or incomplete data. In contrast, \gls{nerf}\cite{nerf} have emerged as a leading implicit representation approach, capable of photorealistic rendering of complex scenes by leveraging neural networks to model scene radiance and density implicitly. Recent advancements such as \gls{3dgs} have further enhanced and optimized implicit representation methods by optimizing \gls{3d} scene representations pipeline, significantly increasing rendering quality and accelerating rendering processes. Despite their impressive fidelity, capturing intricate geometry and illumination details rapidly\cite{instantngp}, it remains computationally intensive, which complicates real-time implementation.

In real-time scenarios such as autonomous robotics and \gls{ar}/\gls{vr}, the timely updating of \gls{3d} scenes is essential, as delays in representation can disrupt system performance and user experience. This concept of timeliness is effectively captured by the metric known as \gls{aoi}, which quantifies data freshness by measuring the time elapsed since the most recent update. Minimizing \gls{aoi} ensures that system decisions are based on the latest information, critical in dynamic environments. However, a practical challenge arises due to inherent network-induced delays, particularly in distributed camera systems, where images from multiple robots or sensors arrive asynchronously at edge servers. This delay leads to a fundamental trade-off: reconstructing scenes quickly using only the freshest images may degrade fidelity, whereas waiting to accumulate additional images can enhance representation quality at the expense of higher \gls{aoi}.

Addressing these intertwined challenges of timeliness and fidelity calls for a holistic design approach that bridges communication and \gls{cv}. Recent developments in 5G and emerging 6G technologies highlight that optimizing communication \glspl{kpi}, such as throughput and latency alone, is insufficient for complex \gls{cv} tasks like \gls{3d} scene representation. Traditional bit-level communication paradigms, rooted in Shannon's theory, consider all transmitted data equally important. In contrast, emerging task-oriented communication frameworks advocate for prioritizing information based on its relevance and semantic significance to the end-task. This paradigm shift underscores the importance of intelligently selecting and transmitting data that significantly enhances the final task outcome, aligning communication strategies directly with \gls{cv} objectives.

Motivated by this paradigm, our research addresses a practical scenario: a wireless network of multiple mobile robot cameras transmitting \gls{2d} images to an edge server tasked with real-time \gls{3d} scene representation. A semantic-driven approach to data transmission is crucial in such settings. Key elements such as camera pose, dynamic objects, and high-salience targets require frequent updates to maintain representation accuracy and system responsiveness. In applications like autonomous driving and mixed reality, dynamic entities such as pedestrians and vehicles must be updated immediately to ensure safety and situational awareness. In contrast, static background components, while updated less frequently, provide essential spatial context for robust scene understanding. Additionally, considering spatial relationships, such as distances and occlusions, further aligns scene representation with practical, real-world applications. Specifically, we focus more on challenging environments such as nuclear decommissioning sites, characterized by complex structures and stringent operational constraints, including limited bandwidth and the critical need for timely and accurate scene updates. These constraints highlight the necessity for efficient, adaptive strategies that optimize the trade-off between \gls{aoi} and representation fidelity.

Integrating semantic criteria into \gls{3d} representation frameworks enables more effective resource prioritization, enhancing fidelity and timeliness specifically for critical scene components. This task-oriented methodology directly aligns with broader advancements in \gls{ai}-driven communication and computation, promoting intelligent, adaptive strategies to meet the stringent demands of emerging real-time applications. Ultimately, this integrated approach supports robust performance in dynamic and safety-critical environments, reflecting a significant evolution beyond traditional communication and representation paradigms.

The key contributions of our work are summarized as follows:
\begin{itemize}
    \item Unified \gls{aoi} and Semantic-Aware Framework: We propose a novel framework integrating data freshness (\gls{aoi}) and semantic relevance, allowing intelligent prioritization of image transmissions based on both temporal and content-based criteria. Semantic importance is evaluated using pre-trained feature extractors, identifying critical scene elements requiring timely updates.
    \item Contextual-Bandit \gls{ppo} Optimization Strategy: We introduce a \gls{rl}-based approach utilizing a contextual-bandit \gls{ppo} algorithm to optimize the selection of incoming image streams dynamically. This method achieves superior fidelity in \gls{3d} scene representation while simultaneously ensuring low \gls{aoi}, and importantly, remains compatible with diverse \gls{nerf} variants, underscoring its broad applicability.
    \item Bridging Communication and Computer Vision: Our study elucidates the interplay between network scheduling policies and their direct impact on \gls{cv} tasks, emphasizing a joint design strategy for optimal performance. Insights gained from this interdisciplinary approach pave the way for future advancements in integrated \gls{ai}-driven communication and vision systems.
\end{itemize}

Through these contributions, our work addresses fundamental gaps in existing methods, proposing a robust solution that significantly enhances real-time \gls{3d} scene representation performance under stringent real-world constraints. This integrated approach underscores the need for a new generation of intelligent, task-oriented designs to meet the rigorous demands of emerging real-time applications.

\section{Related Work}

\subsubsection{Foundations of \gls{3d} Scene Representation}
Explicit \gls{3d} representations such as point clouds~\cite{qi2017pointnet}, voxels~\cite{maturana2015voxnet}, and mesh~\cite{kazhdan2006poisson} are well-established but face scalability and real-time rendering challenges in complex environments. To address these limitations, implicit neural-based methods have emerged, with \gls{nerf}, implicit grids, and their extensions~\cite{rosinol2022nerf,Yu2022MonoSDF,yu2023nerfbridge,Zhu2022CVPR,Sucar:etal:ICCV2021,kerbl20233d} offering superior fidelity and compression by learning continuous volumetric functions. Neural rendering~\cite{rosinol2022nerf} integrates physical projection models with trainable networks, enabling flexible control of lighting, pose, and semantic structure. More recently, \gls{3dgs}~\cite{kerbl20233d} has been proposed as an explicit-yet-efficient representation, using anisotropic Gaussians with rasterization-based blending to achieve high-quality rendering at real-time frame rates. These works establish the foundations for later efforts that adapt \gls{3d} scene modeling to communication-constrained settings.  
\subsubsection{Computer Vision-Oriented Enhancements}
A large body of work in the computer vision community has focused on improving the efficiency and deployability of neural \gls{3d} representations. In~\cite{rosinol2022nerf}, the authors combined dense monocular \gls{slam} with \gls{nerf} to improve pose and depth estimation for real-time scene fitting. In~\cite{Yu2022MonoSDF}, the authors incorporated monocular depth and normal priors to accelerate \gls{nerf} convergence and enhance representation quality. \gls{nerf}-Bridge~\cite{yu2023nerfbridge} linked Nerfstudio~\cite{nerfstudio} with ROS to streamline robotic scene modeling pipelines. Other extensions~\cite{Zhu2022CVPR,Sucar:etal:ICCV2021} advanced \gls{nerf}-based perception in complex settings. Complementarily, efficiency-driven designs such as sparse-data representation~\cite{yadav2025rf,eusebi2024realistic}, model compression~\cite{jia2025towards,yuan2024kv,yang2024video}, and Gaussian splatting improvements support scalability. In~\cite{lin2024rtgsenablingrealtimegaussian}, the authors proposed RTGS, which achieves over 100 FPS Gaussian splatting on mobile devices through pruning and gaze-aware rendering. The authors in~\cite{ren2024octreegsconsistentrealtimerendering} presented Octree-GS, which dynamically selects level-of-detail structures to maintain consistent rendering quality for large-scale scenes. These advances enable implicit and Gaussian-based representations to meet real-time and resource-constrained requirements, providing a technical basis for their integration with communication systems.  

\subsubsection{Integration of Communication and \gls{3d} Scene Representation}
Recent studies have directly incorporated communication efficiency into \gls{3d} scene representation and transmission. In~\cite{zhang20253dgstreaming}, the authors proposed \gls{3dgs} streaming, which combines spatial partitioning, progressive compression, and field-of-view-based bitrate adaptation to reduce latency and improve \gls{qoe} in streaming \gls{3dgs} data. In~\cite{yan2025instant}, the authors introduced Instant Gaussian Stream, employing an anchor-driven Gaussian motion network and key-frame guidance to avoid per-frame optimization, thereby enabling real-time free-viewpoint video transmission. In~\cite{liu2024dynamics}, the authors extended these efforts to 4D streaming with DASS, a dynamics-aware framework that applies selective inheritance, motion-aware shifting, and error-guided densification to improve online representation under communication constraints. Further, the authors in~\cite{liu2025d2gv} investigated Gaussian-to-video conversion for dynamic scene delivery. Beyond visual tasks, the authors in~\cite{wen2025wrf} adapted Gaussian splatting to the radio-frequency domain, introducing WRF-GS for wireless radiation field representation. By integrating an RF projection model with electromagnetic splatting, WRF-GS enables millisecond-level spectrum synthesis and accurate channel state information prediction, surpassing ray tracing and \gls{nerf}-based approaches.  Related works on communication timeliness and freshness also provide complementary insights. Metrics such as \gls{aoi}~\cite{8262774,9322193,9467360}, \gls{voi}~\cite{10013736}, and semantic quality measures such as BLEU~\cite{bleu} and SSIM~\cite{ssim} have been applied in domains including \gls{ugv} control~\cite{10013736}, image transmission~\cite{9959884}, and classification~\cite{9606667}. Agheli et al.~\cite{agheli2023effective} proposed a pull-based sensing architecture with query-driven updates, while OSNeRF~\cite{10757600} designed an on-demand semantic \gls{nerf} framework where a control unit coordinates multi-view data collection by cooperative robots. These works collectively highlight the feasibility of embedding bandwidth, latency, and timeliness constraints directly into \gls{3d} scene modeling and transmission, even without adopting a semantic communication paradigm.  

\subsubsection{Task-Oriented Communications with \gls{3d} Representations}
Building upon the above, recent works explicitly integrate \gls{tosc} with \gls{3d} scene modeling. In~\cite{wu2024semantic}, the authors proposed \gls{nerf}-SeCom for \gls{3d} human face transmission, leveraging a pre-shared \gls{nerf}-based knowledge base and feature selection/prediction to significantly reduce transmission overhead while ensuring visualization quality. In~\cite{yue2025nerfcom}, the authors presented NeRFCom, which employs nonlinear transform coding and \gls{jscc} to enable variable-rate feature transmission and graceful degradation under adverse channel conditions. In~\cite{zhang2024semanticcommunicationrealtime3d}, the authors further demonstrated a semantic communication-based framework where semantic segmentation at the transmitter facilitates efficient \gls{3d} scene representation at the receiver. By explicitly aligning \gls{3d} representations with semantic communication objectives, these works optimize end-to-end fidelity, robustness, and communication efficiency, demonstrating the potential of combining \gls{nerf}/\gls{3dgs} with task-driven semantic transmission paradigms.

\section{System Model}
\begin{figure*}
            \centering
            \includegraphics[width=\textwidth]{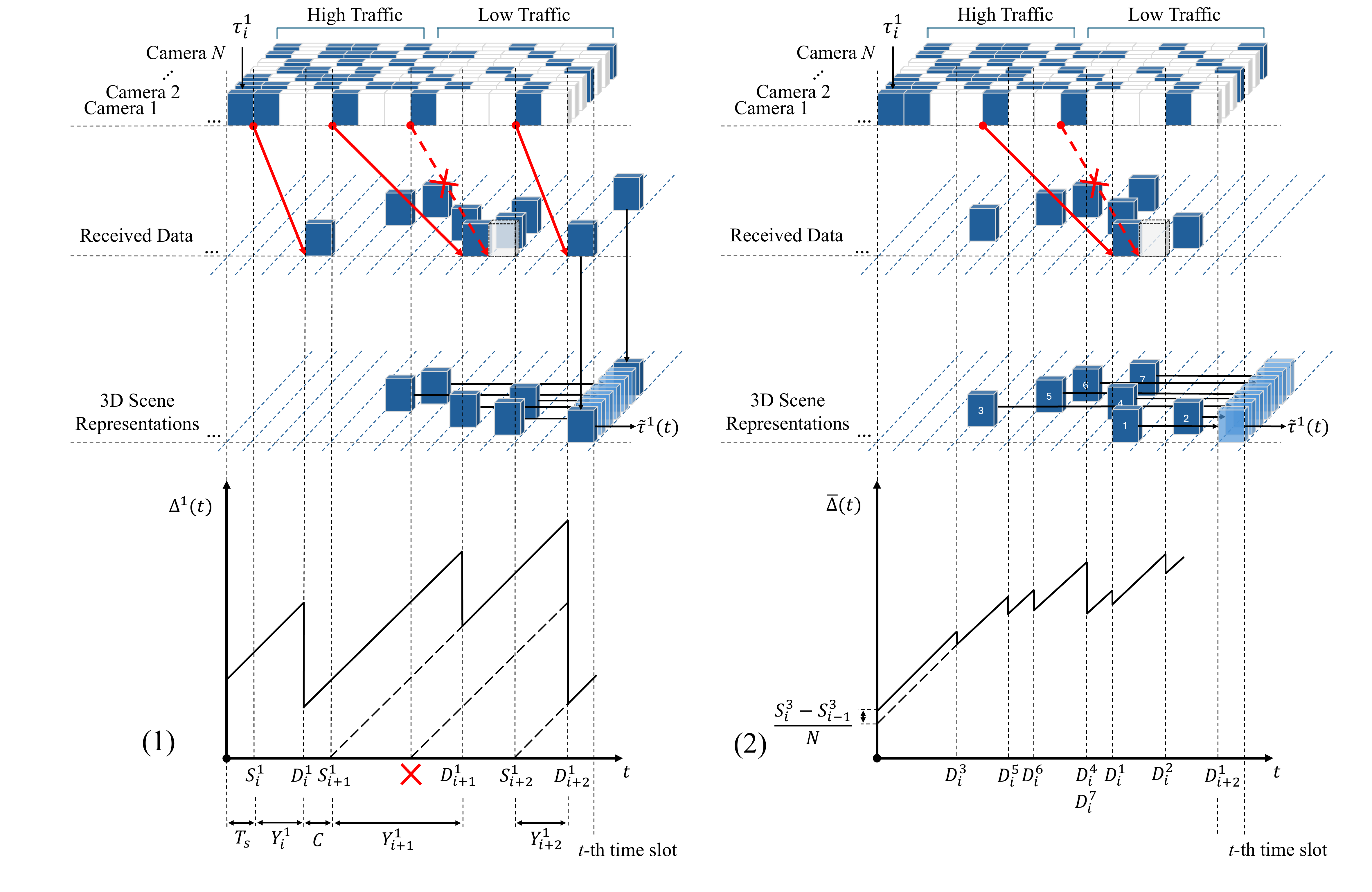}
           \caption{Time-sequence diagrams for real-time dynamic 3D scene representation from multi-sensor image streaming. The figure illustrates two baseline scheduling policies: (1) the $\omega$-threshold policy, where the scheduler includes the most recent images from each camera in the training set only if their current AoI is below a global threshold $\omega_t$, and (2) the $\omega$-wait policy, where the scheduler postpones rendering for $\omega_t$ slots, incorporating only the updates that arrive during this waiting horizon.}
           \label{fig: time}
\end{figure*}
\subsection{Overview}

To investigate the impact of different scheduling strategies on real-time 3D scene representation, we compare two representative time-sequence paradigms under multi-sensor image streaming, namely the $\omega$-threshold policy and the $\omega$-wait policy. In the $\omega$-threshold policy. 
Under the $\omega$-threshold policy, the system utilizes frames whose \gls{aoi} falls within a given threshold prior to the current time slot, thereby reconstructing the scene from relatively fresh but already available data. In contrast, the $\omega$-wait policy defers the reconstruction and waits for new frames to arrive after the current time slot, ensuring that only incoming data is used.
The detailed operation of each policy will be discussed in the following subsections.
Time is discretized into slots, with each slot having a duration of $T_s$. The framework consists of $N$ cameras, indexed by $n = {1,...,N}$, which generate images every $C$ time slots and transmit them to the edge server. To ensure efficient data transmission, we assume that each camera node can determine the state of the channel (idle or busy) using ACK signals and autonomously decide when to generate updates. 
The channel is modeled as a bufferless single-server system with unit capacity, where packet arrivals follow a \gls{mmpp} and the service time is exponentially distributed. The detailed network model will be introduced in Section~\ref{sec:network_model}.
The pose of the $n$-th camera at time $t$ is denoted as
\begin{align}
\mathbf{p}^n(t) = [x^n(t), y^n(t), z^n(t), \theta^n(t), \phi^n(t)] \in \mathbb{R}^{1\times5},
\end{align}
where $[x^n(t), y^n(t), z^n(t)]$ represents the \gls{3d} location, and $[\theta^n(t), \phi^n(t)]$ represents the \gls{2d} viewing direction. The camera poses, which change as robots move, are assumed to be available to the edge server. During the process, each camera generates a sequence of images, with the $i$-th image from the $n$-th camera denoted by $\tau_i^n$, where $i=1,2,...,I$. The transmission of $\tau_i^n$ begins at time slot $S_i^n$ and concludes at $D^n_i$, resulting in a transmission duration of
\begin{align}
Y^n_i = D^n_i - S_i^n,
\end{align}
At time slot $t$, the most recently received image from the $n$-th camera was generated at
\begin{align}
U^n(t) = \max{S^n_i : D^n_i \leq t}.
\end{align}
The \gls{aoi} for the $n$-th camera, as defined in~\cite{6195689}, is
\begin{align}
\Delta^n(t)= t - U^n(t).
\end{align}
In the $t$-th time slot, we define the set of the latest images received at the edge server as
\begin{align}
\widetilde{\mathcal{T}}(t) =[\widetilde{\tau}^1(t), \widetilde{\tau}^2(t),...,\widetilde{\tau}^N(t)].
\end{align}
The camera poses corresponding to those images are defined as
\begin{align}
\mathbf{P}(t) = [{\mathbf{p}}^1(t), {\mathbf{p}}^2(t),...,{\mathbf{p}}^N(t)].
\end{align}
The poses of the cameras used for the \gls{3d} scene representation in the $t$-th time slot are denoted by $\mathbb{P}(t)$, which is a subset of $P(t)$, defined as
\begin{align}
\mathbb{P}(t) &= \{\mathbf{p}^{n_1}(t), \mathbf{p}^{n_2}(t), \ldots, \mathbf{p}^{n_k}(t)\}, \\
\mathbb{I}(t) &= \{\tilde{\mathbf{\tau}}^{n_1}(t), \tilde{\mathbf{\tau}}^{n_2}(t), \ldots, \tilde{\mathbf{\tau}}^{n_k}(t)\}, \\
&\quad n_k \in \{1,\ldots,N_k\}, \ N_ k \le N, \nonumber
\end{align}
where $\mathbb{I}(t)$ is the subset of images corresponding to the selected cameras.

\subsection{Scheduler Agent}
\subsubsection{$\omega$-Threshold policy}

Considering the policy in Fig.~\ref{fig: time}~(1), time is discretized in slots, and for each camera updating time sequence, a blue slot indicates an update and white slots indicate time slots without updates. In the $t$-th time slot, the latest update ${\widetilde{\tau}^1(t)}$ generated by camera 1 at the $(t-2)$-th time slot arrived at the edge server, and the \gls{3d} scene representation is performed with the set of latest images received at the edge server $\widetilde{\mathcal{T}}(t)$. 

Under the $\omega$-threshold policy, the scheduler determines a threshold value $\omega_t$ instead of making individual camera-wise decisions. For the $n$-th camera, its latest received image at time $t$ will be included in the training set if and only if its current \gls{aoi} satisfies
\begin{align}
    \Delta^n(t) < \omega_t .
\end{align}
Accordingly, in the $t$-th time slot, the training set of this scheduling policy is
\begin{align}
    \mathbb{I}_1(t) = \{ \widetilde{\tau}^n(t) \mid \Delta^n(t) < \omega_t, \, 1 \leq n \leq N \}.
\end{align}

\subsubsection{$\omega$-Wait policy}
In the $t$-th time slot, the scheduler decides a waiting horizon $\omega_t$. Instead of immediately performing \gls{3d} scene representation, the system waits until the $(t+\omega_t)$-th time slot. Only updates that arrive strictly between the $t$-th and the $(t+\omega_t)$-th time slot will be used in the \gls{3d} scene representation.  

In the $t$-th time slot, the latest update $\tilde{\tau}^n(t)$ is the $i_n$-th update from the $n$-th camera, hence the training set for the \gls{3d} scene representation of this scheduling policy is 
\begin{align}
\mathbb{I}_2(t) 
= \big\{ \tau^n(t+\omega_t) \;\mid\;& 
Y^n_{i_n+1} - (t - S^n_{i_n+1}) < \omega_t, \\
& 1 \leq n \leq N \big\}.
\end{align}

To evaluate the information freshness under this policy, we adopt the metric of \gls{aaoi}. And the system-wide \gls{aaoi} is defined as
\begin{align}
    \bar{\Delta}(t) = \frac{1}{N}\sum_{n=1}^N {\Delta}^n(t) .
\end{align}


\subsection{Timeliness Embedding Approach}

This approach integrates multiple factors beyond data freshness to determine the relevance of received images for representation. Each image is characterized by:
\begin{equation}
    \mathbf{s}_t = [\Delta^1(t), \dots, \Delta^N(t), \mathbf{p}^1(t), \dots, \mathbf{p}^N(t), \mathbf{z}^1(t), \dots, \mathbf{z}^N(t)],
\end{equation}
where $\Delta^n(t)$ represents the \gls{aoi} of the $n$-th camera, $\mathbf{z}^n(t)$ contains extracted semantic information from the latest update $\tilde{\tau}^n(t)$ of the $n$-th camera.

Let $I(t) \in \mathbb{I}(t)$ denote an image used in the \gls{3d} scene representation in the $t$-th time slot. The impact of an image on \gls{3d} scene representation quality is assessed based on its contribution to scene fidelity, measured using:
\begin{equation}
    M(\mathbf{I}(t), \hat{\mathbf{I}}(t)) = \sum_j w_j \cdot M_j(\mathbf{I}(t), \hat{\mathbf{I}}(t)),
\end{equation}
where $M_j$ includes metrics such as PSNR, SSIM, and LPIPS, $\hat{\mathbf{I}}(t)$ is the novel view synthesis image from the same camera position as image $\mathbf{I}(t)$. The selection of images for representation depends not only on freshness but also on their spatial alignment and contribution to preserving semantic consistency in the scene.

\subsubsection{Weighted Sum Approach}

This approach evaluates the overall tradeoff between \gls{aaoi} and semantic fidelity through a weighted scoring mechanism, prioritizing either timeliness or scene representation quality based on system requirements. The balance between these factors is determined by the weighted function:
\begin{align}
    F_w(t, \omega_t) = w_t \cdot \bar{\Delta}(t) + w_q \cdot Q(t),
\end{align}
where $Q(t)$ quantifies the overall fidelity of the reconstructed scene. The system selects an update strategy that maximizes the tradeoff-adjusted score, which will be detailed in Section~\ref{sec:pf}. Unlike the previous approach, this method directly adjusts the emphasis on timeliness versus fidelity at a global level. The decision whether prioritizing fresher data or higher-quality representations is determined by specific task requirements.

\subsection{Network Model}
\label{sec:network_model}
We model the burstiness of packet arrivals in image transmission using a Markov Modulated Poisson Process (MMPP)~\cite{fischer1993markov}, where network traffic transitions between $M$ states, ${s_0, s_1, ..., s_M}$, governed by a transition matrix. In state $s_m$, packets arrive following a Poisson process with rate $\lambda_g^m$, and transmission delays follow an exponential distribution with rate $\lambda_d^m$:
\begin{equation}
\label{poisson}
X_m^n \sim \text{Poisson}(\lambda_g^m), \quad Y_m^n \sim \text{exp}(\lambda_d^m).
\end{equation}

A simplified model is the Switched Poisson Process (SPP)\cite{hou2018burstiness} with a \gls{ge} channel\cite{gilbert1960capacity}, where the network alternates between low traffic ($G$) and high traffic ($B$) states. The packet generation rate $\lambda_g$ and transmission delay $\lambda_d$ depend on the state:
\begin{equation}
\begin{aligned}
\label{poisson_spp}
& X_i^n \sim \text{Poisson}(\lambda_g), \quad Y_i^n \sim \text{exp}(\lambda_d), \\
& \lambda_g = \lambda_g^L, \quad \lambda_d = \lambda_d^L, \quad \text{if } s=G, \\
& \lambda_g = \lambda_g^H, \quad \lambda_d = \lambda_d^H, \quad \text{if } s=B.
\end{aligned}
\end{equation}

Let matrix $P_t(\Delta t) = e^{Q_s \Delta t}$ denote the probability transition matrix, where $Q_s$ is the generation matrix defined by
\begin{align}\label{qs}
     Q_s &= \begin{bmatrix}
        -\mu_1 & \mu_{12} & \dots & \mu_{1M}\\
        \mu_{21} & -\mu_2 & \dots & \mu_{2M}\\
        \vdots & \vdots & \ddots & \vdots \\ 
        \mu_{M1} & \mu_{M2} & \dots & -\mu_M \\
        \end{bmatrix}, 
\end{align}
where $\mu_{ij}$ denotes the transition rate from state $s_i$ to $s_j$ and $\mu_i= \sum_{j=1, \; j\neq i}^M \mu_{ij}$. For example, if $M=2$, the generator $Q_s$ is given by
\begin{align}
        Q_s & = \begin{bmatrix}
        -\mu_1 & \mu_1\\
        \mu_2 & -\mu_2\\
        \end{bmatrix}, \\
        & = \frac{1}{\mu_1 + \mu_2}
            \begin{bmatrix}
            1 & \mu_1 \\
            1 & -\mu_2
            \end{bmatrix}
            \begin{bmatrix}
            0 & 0 \\
            0 & -(\mu_1+\mu_2)
            \end{bmatrix}\notag
            \begin{bmatrix}
            \mu_2 & \mu_1 \\
            1 & -1
            \end{bmatrix}. \\
\end{align}
Then, the probability transition matrix $P_t(\Delta t)$ of a two-state Markov process is given by
\begin{align}
        P_t(\Delta t) & = \frac{\Delta t}{\mu_1 + \mu_2}
            \begin{bmatrix}
            1 & \mu_1 \\
            1 & -\mu_2
            \end{bmatrix}
            \begin{bmatrix}
            0 & 0 \\
            0 & -(\mu_1+\mu_2)
            \end{bmatrix}\notag
            \begin{bmatrix}
            \mu_2 & \mu_1 \\
            1 & -1
            \end{bmatrix} \\
            & =
            \begin{bmatrix}
                \frac{\mu_2+\mu_1 e^{\Delta t(\mu_1 + \mu_2)}}{\mu_1 + \mu_2} &
                \frac{\mu_1-\mu_1 e^{\Delta t(\mu_1 + \mu_2)}}{\mu_1 + \mu_2} \\
                \frac{\mu_2-\mu_2 e^{\Delta t(\mu_1 + \mu_2)}}{\mu_1 + \mu_2} &
                \frac{\mu_1+\mu_2 e^{\Delta t(\mu_1 + \mu_2)}}{\mu_1 + \mu_2} \\
            \end{bmatrix},
\end{align}
where the entry $p_{i,j}(\Delta t)$ in $P_t(\Delta t)$ represents the probability of being in state $s_j$ after $\Delta t$ time slots, when starting in state $s_i$~\cite{liu2007resource}. 

\subsection{\gls{3d} Scene Representations}
For \gls{3d} scene representations, we employ a representation function 
$\mathcal{F}_\Theta$ parameterized by $\Theta$, which maps camera poses 
$\mathbb{P}(t)$ to the underlying scene representation. 
Depending on the method, $\mathcal{F}_\Theta$ can be instantiated as 
an \gls{mlp}-based \gls{nerf}~\cite{nerf} or an explicit \gls{3d} Gaussian-based model. Specifically, in the $t$-th time slot, $\mathcal{F}_\Theta$ takes poses $\mathbb{P}(t)$ as input and outputs the volume density $\sigma(t)$ and view-dependent RGB color $\textbf{c}(t) = [r, g, b]$, which is expressed by 
\begin{equation}
\{\textbf{c}(t), \sigma(t)\} = {\mathcal{F}_\Theta}\left({\mathbb{P}(t), {\alpha _\Theta }} \right),
\label{eqn: train}
\end{equation}
where the parameters of the neural network are denoted by ${\alpha_\Theta}$. 

To visualize the \gls{3d} scene representations, a \gls{2d} image $\mathbf{\hat{I}}(t)$ can be obtained by volume rendering~\cite{volumerendering}, which is expressed by
\begin{equation}
{\mathbf{\hat{I}}}(t) = {{\cal F}_r}\left( {\textbf{c}(t), \sigma(t),\mathbf{p}_v,\alpha_r} \right),
\label{eqn:render}
\end{equation}
where $\mathbf{p}_v = [x_v, y_v, z_v, \theta_v, \phi_v] \in \mathbb{R}^{1\times5}$ is the pose of the desired view, consisting of the \gls{3d} location $[x_v, y_v, z_v]$ and the \gls{2d} viewing direction $[\theta_v, \phi_v]$. The parameters of the rendering function are denoted by ${\alpha _r}$. Specifically, for the function of volume rendering ${\cal F}_r(\cdot)$, given a ray $\mathbf{r}(q) = \mathbf{o} + q\mathbf{d}, \left|\mathbf{d}\right| = 1$ emanating from the camera position $\mathbf{o}$ and direction $\mathbf{d}$ defined by the specified camera pose and intrinsic parameters.

The representation function $\mathcal{F}_\Theta$ is trained by minimizing 
a representation loss between the predicted outputs of 
$\mathcal{F}_\Theta$ and the ground-truth supervision from images. 
Formally, the training objective is
\begin{align}
\min_\Theta \; \mathcal{L}(\Theta) &= 
\sum_{N_k} M\big(\tilde{\tau}^{n_k}(t), {\mathcal{F}_\Theta}\left({\mathbb{P}(t), {\alpha _\Theta }} \right)\big),\\
n_k &\in \{1,\ldots,N_k\}, \ N_k \le N, \nonumber
\end{align}
where $\ell(\cdot)$ denotes the representation loss measured in image similarity metrics.

\begin{figure}
            \centering
            \includegraphics[width=0.48\textwidth]{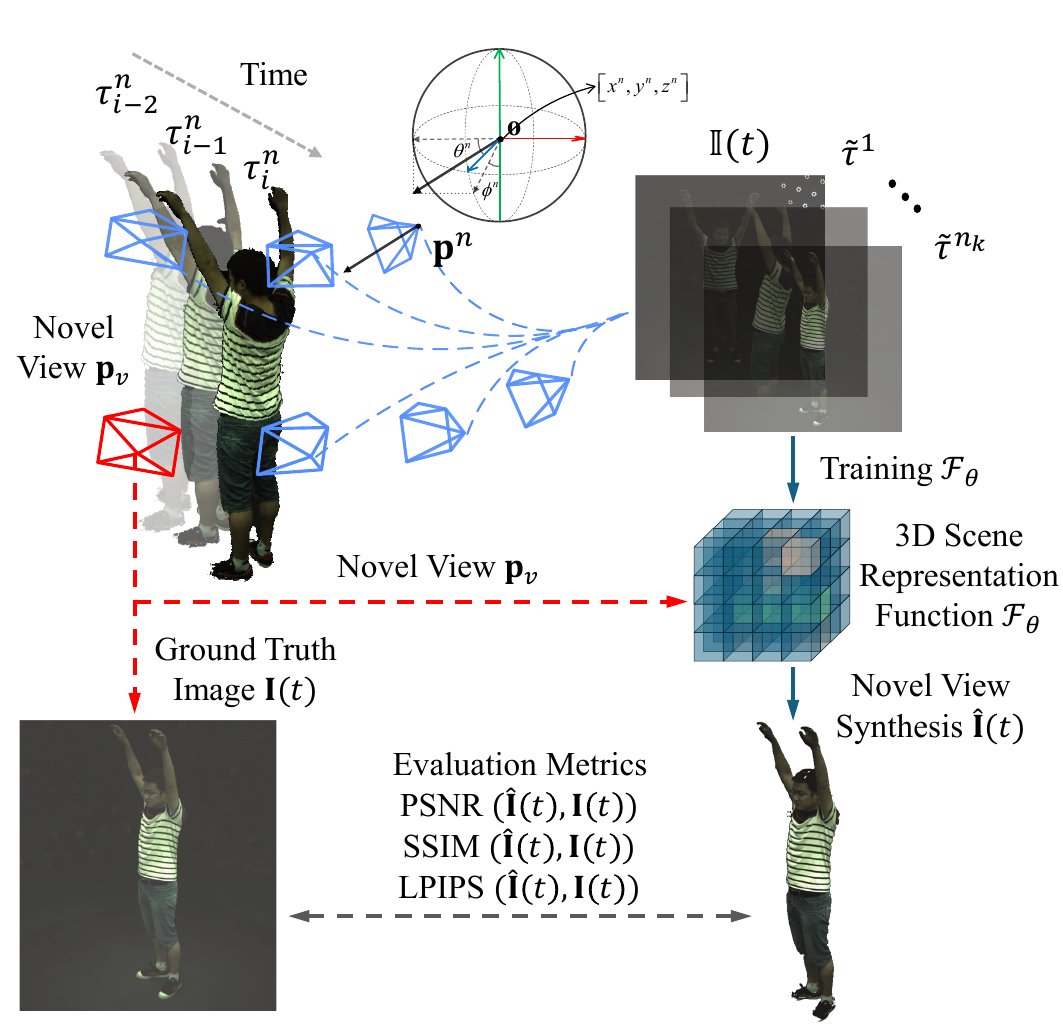}
           \caption{Evaluation pipeline of 3D scene representations. The framework takes the most recent multi-camera images and their corresponding poses $\textbf{p}^n$ as input to a 3D scene representation function $\mathcal{F}_{\Theta}$. Novel view images $\hat{I}(t)$ are synthesized from arbitrary viewpoints $\textbf{p}_v$ using volume rendering. The fidelity of synthesized views is then compared with ground-truth images $I(t)$ through widely used similarity metrics, including PSNR, SSIM, and LPIPS.}
           \label{fig:nerffig}
\end{figure}

\subsection{Task-Oriented Metrics}
To evaluate the performance of \gls{3d} scene representations, we employ the novel view synthesis approach~\cite{fridovich2023k, wang2020unsupervised, dynerf, nerf}. As shown in Fig.~\ref{fig:nerffig}, the novel view synthesis refers to the process of rendering an image from a view that was not originally captured or observed. By comparing different similarity metrics between the reconstructed image and the ground truth, we evaluate the performance of novel view synthesis and the performance of \gls{3d} scene representations from different perspectives~\cite{tewari2022advances}. 

To quantify the performance of novel view synthesis, we adopt three widely used metrics, i.e., \gls{psnr}, \gls{ssim}~\cite{ssim}, and \gls{lpips}~\cite{lpips}. 
In the $t$-th time slot, the \gls{psnr} metric evaluates the per-pixel level picture fidelity and accuracy by comparing the ground truth of the image $\mathbf{I}(t)$ and the synthesized image $\mathbf{\hat{I}}(t)$ from \gls{3d} scene representations, 
\begin{equation}
    \text{PSNR}\left(\mathbf{I}(t), \mathbf{\hat{I}}(t)\right) = 10 \cdot \log_{10}\left(\frac{R_I^2}{\text{MSE}(\mathbf{I}(t), \mathbf{\hat{I}}(t))}\right),
    \label{psnr1}
\end{equation}
\begin{equation}
        \text{MSE}\left(\mathbf{I}(t), \mathbf{\hat{I}}(t)\right) = \frac{1}{L_H L_W}\sum_{m=1}^{L_H}\sum_{n=1}^{L_W}\left(i_{m, n}(t)-\hat{i}_{m, n}(t)\right)^2, 
\end{equation}
where $\mathbf{I}(t)$ is the image of the ground truth of the novel view, and $\mathbf{\hat{I}}(t)$ is the synthesized image from \gls{3d} scene representations. $i_{m, n}(t)$ and ${\hat{i}}_{m, n}(t)$ are the $m$-th row by $n$-th column pixel value of the given image $\mathbf{I}(t)$ and $\mathbf{\hat{I}}(t)$, respectively. The height and width pixel numbers of the images are denoted by $L_H$ and $L_W$, respectively, and $R_I = 2^{\kappa}-1$ is the maximum fluctuation for an image of $\kappa$-bit color per pixel. 

To evaluate the difference in luminance, contrast, and structural information, we utilize \gls{ssim}, 
\begin{equation}
\begin{aligned}
       &\text{SSIM}\left(\mathbf{I}(t), \mathbf{\hat{I}}(t)\right)
       \\
      &\quad\; =\frac{\left(2\mu_\mathbf{I}\mu_\mathbf{\hat{I}}+(k_1L_d)^2\right) \left((2\sigma_{c}+(k_2L_d)^2\right) }{\left(\mu_\mathbf{I}^2+\mu_\mathbf{\hat{I}}^2+(k_1L_d)^2\right)\left(\sigma_\mathbf{I}^2+\sigma_\mathbf{\hat{I}}^2+(k_2L_d)^2\right)},
\end{aligned}
\end{equation}
where the average pixel value of $\mathbf{I}(t)$ and $\mathbf{\hat{I}}(t)$ are denoted by $\mu_\mathbf{I}$ and $\mu_\mathbf{\hat{I}}$, respectively, the standard deviations of the pixel values of $\mathbf{I}(t)$ and $\mathbf{\hat{I}}(t)$ are denoted by $\sigma_\mathbf{I}$ and $\sigma_\mathbf{\hat{I}}$, respectively, the covariance between $\mathbf{I}(t)$ and $\mathbf{\hat{I}}(t)$ is $\sigma_{c}$, $k_1$ and $k_2$ are the parameters to avoid instability, and the dynamic range of the images is denoted by $L_d$. 

To evaluate the perceptual similarity between images, a learning-based perceptual image patch similarity metric, \gls{lpips}, is also used. The idea is to estimate the visual similarity of human perception by learning a neural network model. The model uses a convolutional neural network (CNN) to perform feature extraction on local patches of an image and calculates the similarity score between patches~\cite{lpips}. In the $t$-th time slot, the \gls{lpips} is
\begin{equation}
    \text{LPIPS}\left(\mathbf{I}(t), \mathbf{\hat{I}}(t)\right) = \mathcal{G}\left(d\left(\mathbf{I}(t), \mathbf{\hat{I}}(t)\right)\right),
    \label{lpips1}
\end{equation}
\begin{equation}
    d\left(\mathbf{I}(t), \mathbf{\hat{I}}(t)\right) = \sum^{L}_{l=1} \frac{1}{L_H L_W} {\lVert w_l \odot \left(\mathbf{I}'_l(t) - \mathbf{\hat{I}}'_l(t)\right) \rVert}_2^2,
    \label{lpips2}
\end{equation}
where $\mathbf{I}'_l(t)$ and $\mathbf{\hat{I}}'_l(t)$ are the features extracted at the $l$-th layer of the neural network with input $\mathbf{I}(t)$ and $\mathbf{\hat{I}}(t)$, respectively. The activation functions for each layer of the neural network are denoted by vector $w_l$. $\mathcal{G}$ is the neural network that predicts the \gls{lpips} metric value from input distance $d(\mathbf{I},\mathbf{\hat{I}})$. ${\lVert \cdot \rVert}_2$ is the $\ell_2$ norm distance and $\odot$ denotes the tensor product operation.

\section{Problem Formulation}\label{sec:pf}
To achieve optimal fidelity \gls{3d} scene representation results, we proposed a \gls{drl} method that leverages the \gls{aoi} and semantics from images and camera poses to decide if the received image needs to be involved in \gls{3d} scene representations and rendering to optimize the performance of novel view synthesis. Specifically, we propose to use the \gls{ppo} algorithm as the baseline method for its simplicity, effectiveness, and high sample efficiency~\cite{ppo}. Since our framework encompasses only the randomness induced by the one-step dynamics of the environment, we modify the standard \gls{ppo} algorithm to a single-step contextual-bandit \gls{ppo} to solve the problem~\cite{singlestep}.




\subsection{State}
In the $t$-th time slot, let $\tilde{\tau}^n(t)$ denote the most recent image captured by the $n$-th camera.  
A semantic feature extractor $\mathcal{F}_e(\cdot)$ is employed to encode the image $\Tilde{\tau}^n(t)$ into a compact feature embedding:  
\begin{align}
\mathbf{z}^n_t = \mathcal{F}_e(\Tilde{\tau}^n(t)), \quad \mathbf{z}^n_t\in \mathbb{R}^{1 \times d},
\end{align}
where $d$ is the feature dimension determined by the extractor architecture. The semantic feature extractor $\mathcal{F}_e(\cdot)$ is implemented as the YOLOv11 backbone with default pretrained parameters~\cite{yolo11_ultralytics}.

The state $\mathbf{s}_t$ in the $t$-th time slot is constructed by combining the \gls{aoi} values, semantic feature vectors, and positions of all $N$ cameras.  
Let 
\begin{align}
Z_t = \begin{bmatrix}\mathbf{z}^1_t \\[-2pt] \vdots \\[-2pt] \mathbf{z}^N_t \end{bmatrix} \in \mathbb{R}&^{N \times d},\quad
\boldsymbol{\Delta}(t) = \begin{bmatrix}\Delta^1(t) \\[-2pt] \vdots \\[-2pt] \Delta^N(t)\end{bmatrix} \in \mathbb{R}^{N \times 1},\quad \\
\mathbf{P}(t) = &\begin{bmatrix}\mathbf{p}^1(t) \\[-2pt] \vdots \\[-2pt] \mathbf{p}^N(t)\end{bmatrix} \in \mathbb{R}^{N \times 5}.
\end{align}

where $\mathbf{z}^n_t$ is the semantic feature embedding of the $n$-th camera, $\Delta^n(t)$ is its \gls{aoi} at time $t$, and $\mathbf{p}^n(t)$ denotes its position vector.  
By concatenating these components, we obtain
\begin{align}
\mathbf{s}_t = \big[ Z_t \ \ \boldsymbol{\Delta}(t) \ \ P_t \big] \in \mathbb{R}^{N \times (d+6)}.
\end{align}

\subsection{Action}
In the $t$-th time slot, the scheduling agent selects an action corresponding to $\omega_t$, defined as 
\begin{align}
\mathbf{a}_t = \omega_t \in \{0,1,\ldots, \omega_{\max}\},
\end{align}
where $\omega_t=0$ indicates immediate rendering with the currently available images, and $\omega_t>0$ specifies that the scheduler wait for $\omega_t$ time slots to incorporate potentially fresher updates.




\subsection{Reward}
The tradeoff is quantified through the weighted function, 
\begin{align}
\label{fwr}
    F_w(t, \omega_t) = w_t \cdot \bar{\Delta}(t) + w_q \cdot Q(t),
\end{align}
where the fidelity qualifier $Q(t)$ is measured as a weighted combination of the three image similarity metrics:
\begin{align}
     Q(t) &= w_p \times\text{PSNR}(\mathbf{I}(t), \mathbf{\hat{I}}(t)) \nonumber+ w_s \times \text{SSIM}(\mathbf{I}(t), \mathbf{\hat{I}}(t))\\  &+ w_l \times \text{LPIPS}(\mathbf{I}(t), \mathbf{\hat{I}}(t)).
\end{align}
For compactness, let $w_1 = w_t\times w_p$,  $w_2 = w_t\times w_s$ and $w_3 = w_t\times w_l$, the function in eq.~\ref{fwr} can then be rewritten as
\begin{align}
    F_w(t, \omega_t) &= w_1 \times\text{PSNR}(\mathbf{I}(t), \mathbf{\hat{I}}(t)) \nonumber+ w_2 \times \text{SSIM}(\mathbf{I}(t), \mathbf{\hat{I}}(t))\\  &+ w_3 \times \text{LPIPS}(\mathbf{I}(t), \mathbf{\hat{I}}(t)) + w_t \times \bar{\Delta}(t),
\end{align}
which is a weighted sum of the three metrics we introduced in the last section, where the timeliness–fidelity tradeoff performance is monotonically non-increasing in $F_w(t, \omega_t)$.

Given the state $\mathbf{s}_t$ and action $\mathbf{a}_t$ in the $t$-th time slot, the instantaneous reward in the $t$-th time slot, is defined as 
\begin{equation}
\begin{aligned}
    r(\mathbf{s}_t, \mathbf{a}_t)  = -F_w(t, \omega_t),
\end{aligned}
\end{equation}
where the negative sign ensures that maximizing the instantaneous reward $r(\mathbf{s}_t, \mathbf{a}_t)$ is equivalent to minimizing the tradeoff weighted function $F_w(t, \omega_t)$.
Depending on the application scenario, we can set $w_1$, $w_2$, $w_3$, and $w_t$ to be to different values.


\begin{algorithm}[t]
\caption{Contextual PPO}
\label{algorithm1}
\begin{algorithmic}[1]
\State \text{Input}: Initialize the parameters of the channel model with communication delay distribution parameter $\lambda$, initial parameters of the neural network $\theta_0$, novel view pose $\mathbf{p}_v$, training steps $T_t$, parameters $\alpha_\Theta$ of the representation function $\mathcal{F}_\Theta$. 
\For{$t = 1, 2, ... T_t$}
    \State $\mathbf{s}_t \gets [\Delta^1(t), \Delta^2(t),..., \Delta^N(t)]$ Obtain the state from 
    \Statex \quad \; AoIs.
    \State $\mathbf{a}_t \gets \pi_{\theta_t}(\mathbf{s}_t)$ Take the action $\mathbf{a}_t$  from $\pi_{\theta_t}(\mathbf{s}_t)$.
    \State $\mathbf{\hat{I}}(t) \gets$ Train $\mathcal{F}_\Theta$ and render the image from novel
    \Statex \quad \; view $\mathbf{p}_v$.
    \State $r(\mathbf{s}_t, \mathbf{a}_t) \gets \text{LPIPS}(\mathbf{I}(t), \mathbf{\hat{I}}(t)), \text{SSIM}(\mathbf{I}(t), \mathbf{\hat{I}}(t)), \bar{\Delta}(t), $ 
    \Statex \quad \; $\text{PSNR}(\mathbf{I}(t), \mathbf{\hat{I}}(t))$~ Calculate the instantaneous reward 
    \Statex \quad \; by (\ref{psnr1})-(\ref{lpips2}).
    \State $A^{\pi_\theta} \gets \; Q^{\pi_{\theta_t}} (\textbf{s}_t, \textbf{a}_t) -  V^{\pi_{\theta_t}}(\textbf{s}_t)$~Calculate advantage 
    \Statex \quad \; with~(\ref{qvalue}), (\ref{value}).
    \State $\pi_{\theta_{t+1}} \gets A^{\pi_\theta}$ Take one-step policy update towards 
    \Statex \quad \; maximizing $\mathcal{L}(\textbf{s}_t, \textbf{a}_t, \theta_t, \theta)$ in~(\ref{loss}).
\EndFor
\State \textbf{Output}: Optimal policy $\pi^*_{\theta}$.
\end{algorithmic}
\end{algorithm}

\subsection{Problem Formulation}
The policy $\pi_\theta$ maps the state $\mathbf{s}_t$ to a categorical distribution over the waiting actions.  
For the $n$-th camera, the action $a_t^n$ represents the selected $\omega$-wait duration from the discrete set 
$\{0,1,\ldots,\omega_{\max}\}$.  
The policy outputs the probability distribution
\begin{align}\label{eq:sampling}
\bm{\rho}_t^n \triangleq
\begin{pmatrix}
   \Pr\{a_t^n=0\} \\
   \Pr\{a_t^n=1\} \\
   \vdots \\
   \Pr\{a_t^n=\omega_{\max}\}
\end{pmatrix} \in \mathbb{R}^{(\omega_{\max}+1) \times 1},
\end{align}
where $\bm{\rho}_t^n$ is the categorical distribution over all possible waiting times for the $n$-th camera at slot $t$.

The policy is represented by a neural network denoted by $\pi_{\theta}(\mathbf{s}_t)$
, where $\theta$ are the training parameters.
Following the $\pi_\theta$ policy, the long-term reward is given by
\begin{equation}
    R^{\pi_\theta} = \mathbb{E}[\sum_{t=0}^{\infty} \gamma^t r(\mathbf{s}_t, \mathbf{a}_t)],
\end{equation}
where $\gamma$ is the reward discounting factor. To find an optimal policy $\pi_\theta^*$ that maximizes the long-term reward $R^{\pi_\theta}$, the problem is formulated as
\begin{align}
    &{\pi_{\theta}^{*}}  =  \mathop {\max }\limits_{\theta} Q^{\pi_{\theta}}({\bf{s}}_t, {\bf{a}}_t)\hfill,    \\
    \label{qvalue}
    Q^{\pi_{\theta}}({\bf{s}_t}, {\bf{a}_t}) = & \;{\mathop{\mathbb{E}}}[\sum_{t = 0}^\infty  {{ \gamma ^t}}r({\bf{s}}_t, {\bf{a}}_t) \mid {\bf{s}}_0={\bf{s}},\ {\bf{a}}_0={\bf{a}},\ \pi_{\theta}],
\end{align}
where $Q^{\pi_\theta}$ is the state-action value function.

\section{Contextual-bandit \gls{ppo} Algorithm}
The \gls{ppo} algorithm updates the parameters of the policy neural network $\theta_t$ by
\begin{align}\label{loss}
\mathcal{L}({\mathbf{s}}_t,{\mathbf{a}}_t &,\theta_t,\theta) = \min \left(\frac{\pi_{\theta}({\mathbf{a}}_t \mid {\mathbf{s}}_t)}{\pi_{\theta_{t}}({\mathbf{a}}_t \mid {\mathbf{s}}_t)}{A^{\pi_{\theta_t}}({\mathbf{s}}_t,{\mathbf{a}}_t)},\right. \\ \notag
&\left.\text{clip}\left(\frac{{\pi_\theta}({\bf{a}}_t \mid {\bf{s}}_t)}{\pi_{\theta_t}({\bf{a}}_t \mid {\bf{s}}_t)}, 1-\epsilon, 1+\epsilon,\right)A^{\pi_{\theta_t}}({\bf{s}}_t,{\bf{a}}_t)\right),
\end{align} 
where $A^{\pi_{\theta_t}}$ is the advantage function estimating the advantage of taking action $\bf{a}_t$ in state $\bf{s}_t$~\cite{gae}, which is expressed by
\begin{align}
\label{advantage}
A^{\pi_{\theta_t}} = &\; Q^{\pi_{\theta_t}}(\textbf{s}_t, \textbf{a}_t) -  V^{\pi_{\theta_t}}(\textbf{s}_t),\\
\label{value}
V^{\pi_{\theta_t}}({\bf{s}}_t) = &\; {\mathop{\mathbb{E}}}[\sum_{t = 0}^\infty  {{ \gamma^t}}r({\bf{s}}_t, {\bf{a}}_t) \mid {\bf{s}}_0={\bf{s}},\ \pi_{\theta}], 
\end{align}
where $V^{\pi_{\theta_t}}$ is the state-value function.

The details of the proposed contextual-bandit \gls{ppo} algorithm are shown in~\cref{algorithm1}. First, different parameters of neural networks and channel models are initialized. In the $t$-th time slot, based on the \glspl{aoi} of each camera, and the semantic information from images and camera poses, we obtain the state, $\mathbf{s}_t$, and take the action, $\mathbf{a}_t$. Then, by calculating the weighted sum reward from the \gls{3d} scene representation and rendering results, we obtain the instantaneous reward. After that, we estimate the advantage $A(\mathbf{s}_t,\mathbf{a}_t)$ of taking action $\mathbf{a}_t$ in state $\mathbf{s}_t$ with the state-action value function (\ref{qvalue}) and state-value function (\ref{value}). With the advantage $A(\mathbf{s}_t,\mathbf{a}_t)$ we take one step gradient descend to update the current policy $\pi_{\theta_t}$ to $\pi_{\theta_{t+1}}$ by maximizing the \gls{ppo} loss function (\ref{loss}).



\section{Experiment Setup}
\subsection{Evaluation of 3D Scene Representations}

To comprehensively evaluate the performance of \gls{3d} scene representations, we consider both controlled benchmarks and realistic communication scenarios.  
The DyNeRF dataset provides high-quality inward-facing multi-view sequences for assessing representation accuracy and timeliness, while the Eyeful Tower dataset introduces large-scale outward-facing scenes with greater environmental complexity.  
In addition, we examine the influence of network burstiness using a Gilbert–Elliott channel model to capture the impact of packet-level dynamics on real-time scene representation.  
In our configuration, the packet generation rate and transmission delay follow the state-dependent parameters 
$\lambda_g^H = 1/30$, $\lambda_g^L = 1/120$, $\lambda_d^H = 1/60$, and $\lambda_d^L = 1/30$.  
The underlying Markov chain is a two-state process ($M=2$), where the transition rate $\mu_1 = \mu_2 = 1/30$.  
This configuration captures the contrast between high-state burstiness with longer delays and low-state sparsity with shorter delays.

\begin{figure*}
            \centering
            \includegraphics[width=\textwidth]{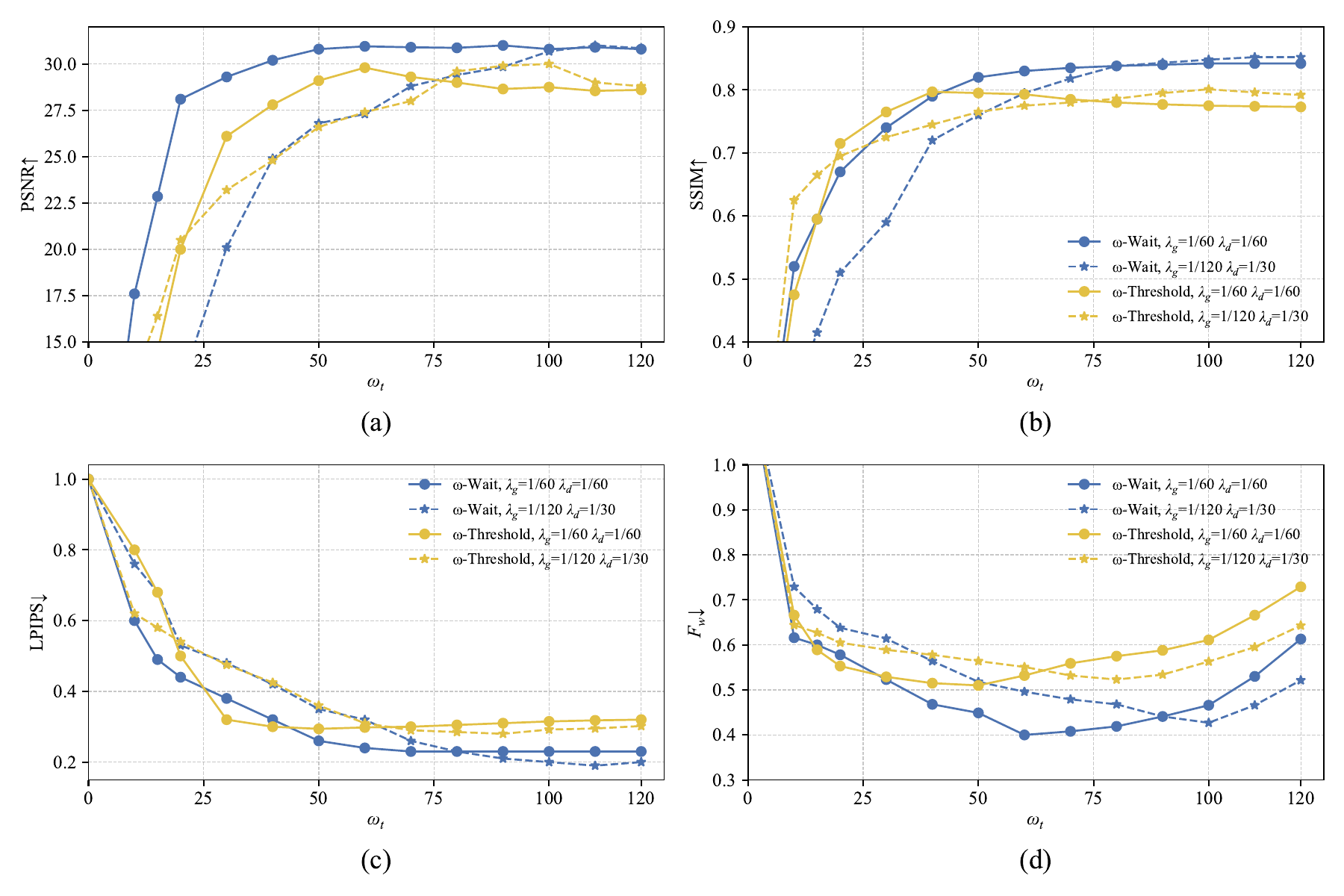}
           \caption{Comparison of representation quality and overall performance under different scheduling strategies averaged along different datasets with Instant-NGP.
           The four subfigures report (a) PSNR↑, (b) SSIM↑, (c) LPIPS↓, and (d) the $F_w$↓ as functions of the parameter $\omega_t$. Results are shown for two traffic intensities ($\lambda_g$=1/60, $\lambda_d=1/60$ and $\lambda_g$=1/120, $\lambda_d=1/30$) and two policies ($\omega$-wait and $\omega$-threshold). 
           }
           \label{fig:big_res}
\end{figure*}

\subsection{Evaluation on the DyNeRF and the ZJU-Mocap Dataset}

The DyNeRF dataset~\cite{dynerf} is widely used for \gls{3d} scene representation and novel view synthesis. It consists of 10-second, 30-FPS multi-view videos captured from 19 cameras positioned at different angles, providing high-quality data for training and evaluation. The ZJU-MoCap dataset~\cite{zjumocap} contains human motion sequences recorded in a multi-view studio setup with synchronized cameras. Both datasets adopt inward-facing camera configurations, which are particularly suited for neural rendering methods that rely on dense multi-view observations.

For benchmarking, we adopt the following methodology:
\begin{itemize}
    \item Training and evaluation: Eighteen out of the 19 available videos are used to train the \gls{3d} scene representations, while the remaining one serves as the ground truth for evaluating novel view synthesis performance, i.e., N=18.
    \item Frame capture interval: Each camera captures frames at a fixed interval of 30 ms.
\end{itemize}

\begin{figure}
    \centering
    \includegraphics[width=0.48\textwidth]{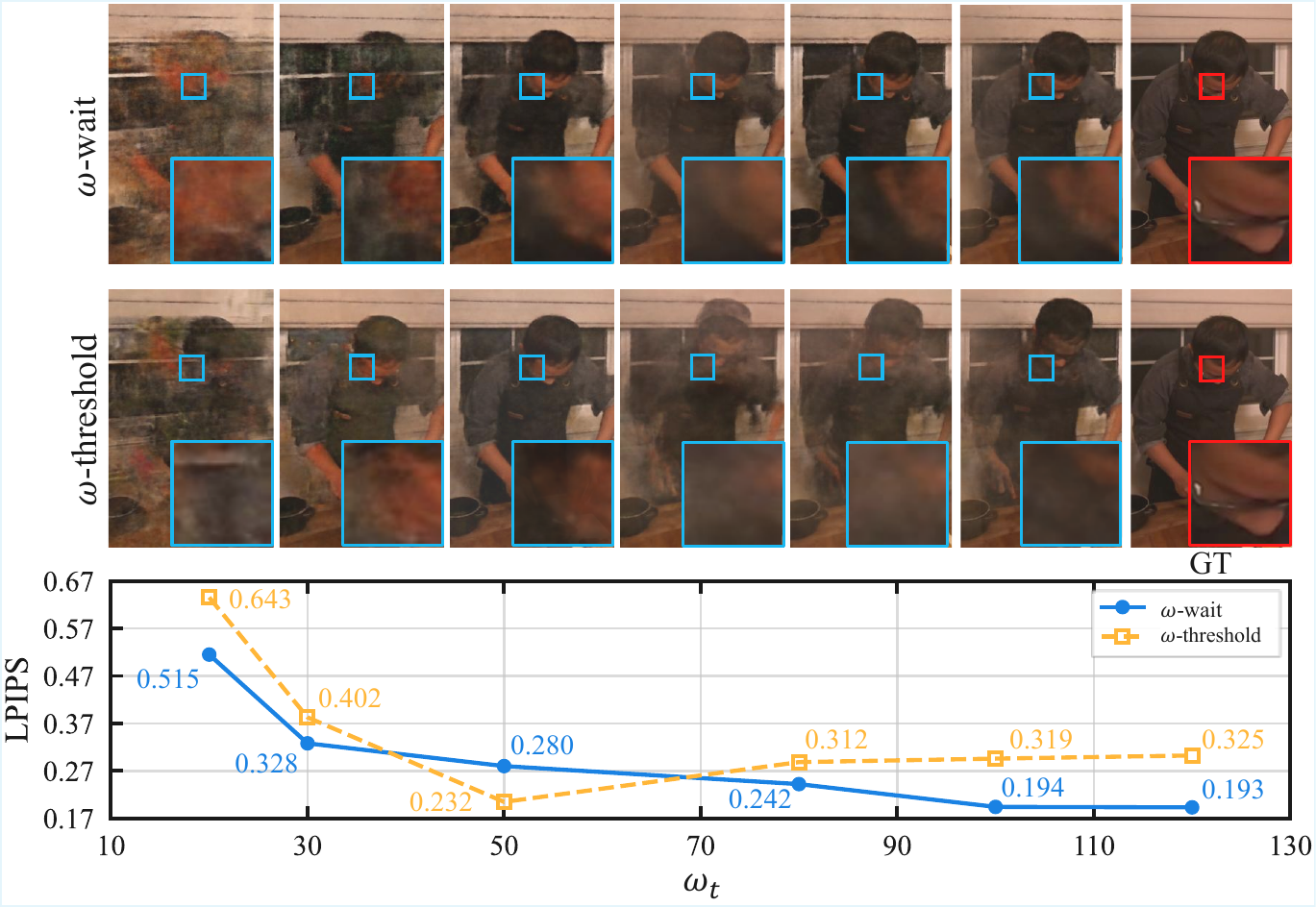}
    \caption{Comparison of \gls{3d} scene representation quality under the $\omega$-wait and the $\omega$-threshold policy, trained with Instant-NGP.}
    \label{fig:results_compare}
\end{figure}

\begin{figure}
    \centering
    \includegraphics[width=0.485\textwidth]{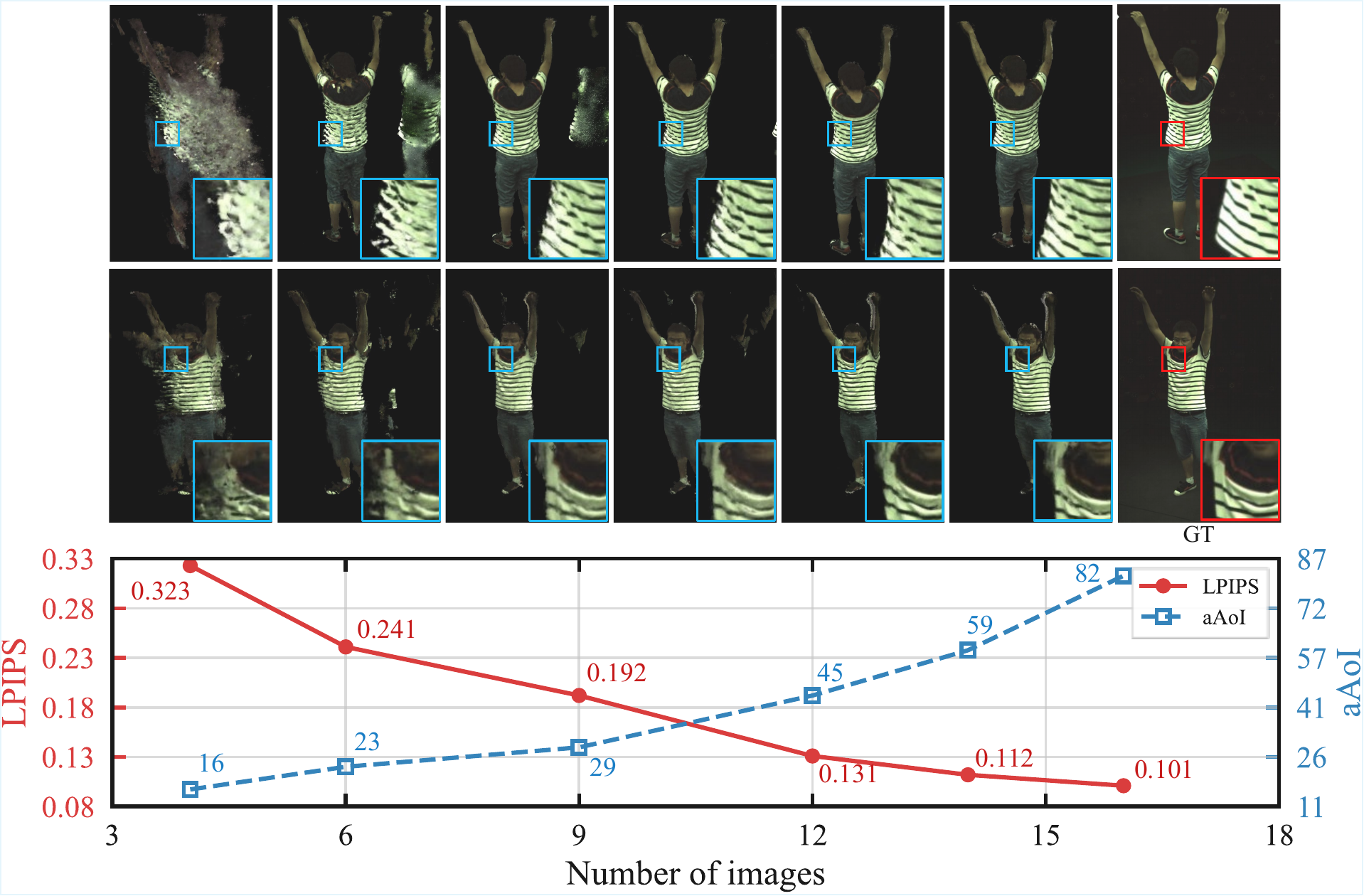}
    \caption{Results on the ZJU-MoCap dataset, trained with Nerfacto. The plots report LPIPS↓ and aAoI↑ as the number of training images increases.}
    \label{fig:result_4}
\end{figure}
\begin{figure}
    \centering
    \includegraphics[width=0.485\textwidth]{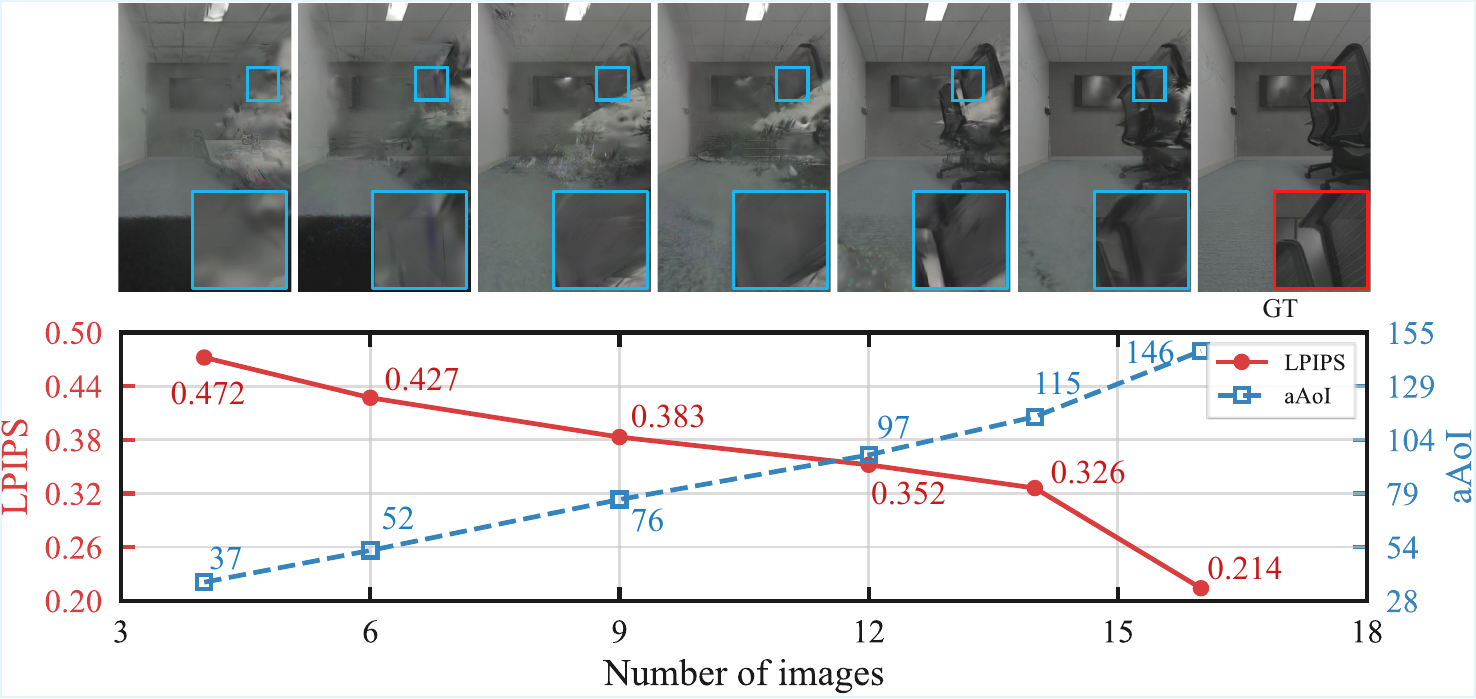}
    \caption{Results on the VR-NeRF Eyeful Tower dataset, trained with 3D Gaussian Splatting. The plots report LPIPS↓ and AoI↑ as the number of training images increases.}
    \label{fig:results_5}
\end{figure}

\subsection{Evaluation on the VR-NeRF Eyeful Tower Dataset}

The VR-NeRF Eyeful Tower dataset~\cite{VRNeRF} is an outward-facing multi-view dataset specifically designed for evaluating large-scale \gls{3d} scene representation and novel view synthesis.  
Unlike inward-facing datasets such as DyNeRF, the cameras in Eyeful Tower are positioned around outdoor landmarks and point outward, capturing diverse views of the scene under natural illumination.  
This setting introduces greater challenges in terms of scale variation, occlusion, and background complexity, making it suitable for benchmarking semantic-aware scheduling policies.

For benchmarking, we adopt the following methodology:
\begin{itemize}
    \item Training and evaluation: A subset (N = 18) of the outward-facing camera views is used for training the \gls{3d} scene representations, while the remaining views are held out for evaluating novel view synthesis performance.
    \item Frame capture interval: Each camera records frames at a fixed interval of 33 ms (corresponding to 30 FPS).
\end{itemize}




\subsection{Evaluation on Packet Burstiness}
To further investigate the impact of communication dynamics on \gls{3d} scene representations, we simulate a \gls{ge} channel model, which introduces packet burstiness and network latency into the pipeline. Before the simulation begins, a time series map is generated for each of the 19 sensor nodes, containing both the sending and receiving frame sequences. At each packet interval \(T_i\), a frame is sampled and transmitted. After experiencing the total uplink delay \(Y_i^n\), the frame is stored in the training buffer of the corresponding node within the scene representation module.

\subsection{\gls{3d} Scene Representation Methods}
To analyze the tradeoff between timeliness and fidelity in real-time \gls{3d} scene representations, we evaluate our approach using three well-known \gls{nerf}-based methods:

\subsubsection{\gls{ngp}~\normalfont{\cite{instantngp}}}
\gls{ngp} is a \gls{nerf}-based \gls{3d} scene representation method. Utilizing a C++ embedded neural network structure, hash coding, and \gls{cuda}, \gls{ngp} achieves state-of-the-art training speed among all \gls{nerf} methods.

\subsubsection{\gls{3d} Gaussian Splatting \normalfont{\cite{kerbl20233d}}}
\gls{3dgs} represents a scene using a set of anisotropic \gls{3d} Gaussians instead of an implicit neural field.  
Each Gaussian is parameterized by its position, covariance, opacity, and color, and the scene rendering is achieved by splatting these Gaussians along camera rays.  
This explicit representation enables real-time training and rendering, offering a significant speedup compared to traditional \gls{nerf}-based methods while preserving high visual fidelity.

\subsubsection{Nerfacto \normalfont{\cite{nerfstudio}}}
Nerfacto integrates camera pose refinement and per-image appearance conditioning to augment representation quality. It applies the hash coding from \gls{ngp} to accelerate training. Compared with \gls{ngp}'s \gls{cuda}-based core computing module, Nerfstudio is programmed in Python and thus needs more computation time.

\section{Performance Evaluation}
In~\ref{timeliness}, to demonstrate the timeliness–fidelity tradeoff, we consider both the $\omega$-threshold and the $\omega$-wait methods. In~\ref{rl}, to show the optimal performance achieved by the proposed contextual-bandit \gls{ppo} algorithm, we evaluate the training and testing results. The parameters of the $F_w(t, \omega_t)$ are set to $w_1 = -0.02, w_2 = -0.2, w_3 = 0.3, w_t = 0.015$.

\subsection{Timeliness-Fidelity Tradeoff with $\omega$-Threshold Policy}\label{timeliness}

Figure~\ref {fig:big_res} illustrates the effect of the scheduling parameter $\omega_t \in (0, 120]$ in the $\omega$-Threshold policy on the performance of \gls{3d} scene representations, measured by the three image similarity metrics and $F_w$, respectively. The results highlight an inherent tradeoff between timeliness and fidelity. In particular, as $\omega_t$ increases, the system tends to wait for more packets, improving delivery rates but also raising the risk of incorporating outdated frames. Consequently, when $\omega_t$ becomes excessively large, representation quality degrades due to the contamination of stale information.  

In the high-traffic state ($\lambda_g=\lambda_d=1/60$), \gls{3d} scene representation quality is initially low because of high latency, yet it improves rapidly as packet delivery rate increases. The trade-off point is reached earlier at $\omega_t = 50$.
The corresponding $F_w$ curve exhibits a sharp decline at small $\omega_t$, capturing the improvement in timeliness, but then rises once the inclusion of stale frames outweighs the benefit of higher delivery rates. The low-traffic state ($\lambda_g=1/120, \lambda_d=1/30$) starts from a higher fidelity owing to reduced latency, but the growth of representation quality is slower due to lower packet arrival rates. Nevertheless, the tradeoff point is higher at $\omega_t = 72$.
Since packet drops in the queue are less frequent, the probability of outdated frame contamination is reduced. Overall, these results confirm that the choice of $\omega_t$ naturally induces a tradeoff in image metrics alone.



\begin{figure*}
            \includegraphics[width=\textwidth]{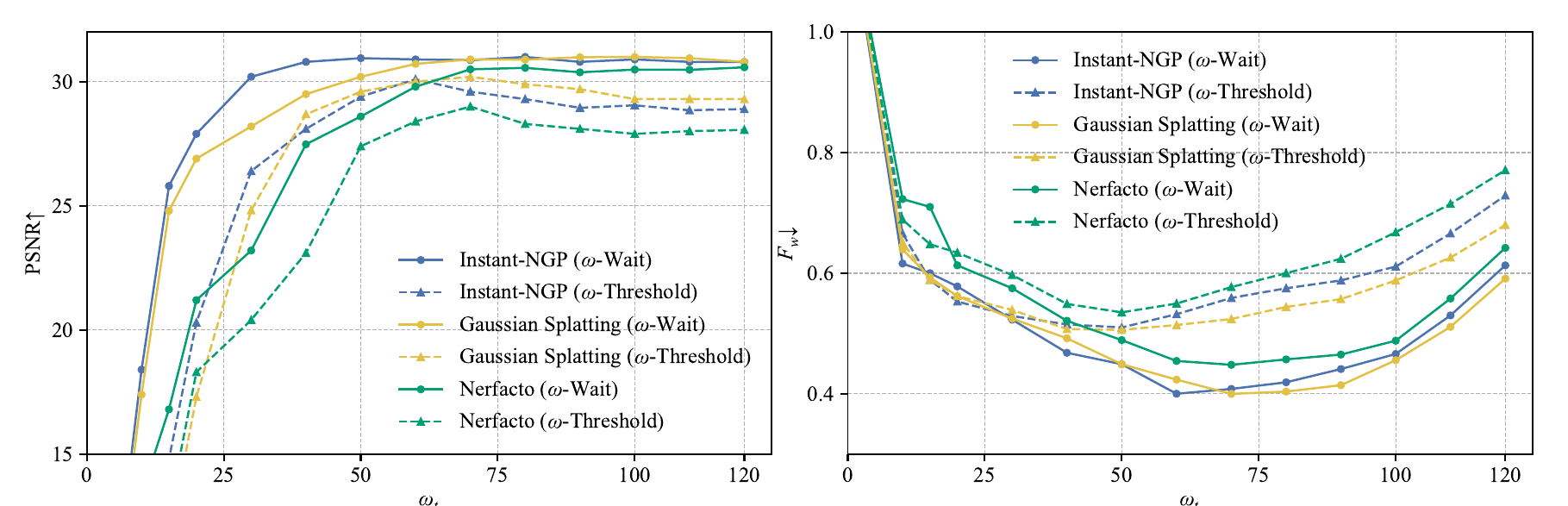}
           \caption{Comparison of representation quality and overall performance under different scheduling strategies and \gls{3d} scene representation methods. 
           The two subfigures report PSNR↑ and the $F_w$↓ as functions of the parameter $\omega$. Results are shown for two policies ($\omega$-wait and $\omega$-threshold). 
           }
           \label{fig:methods}
\end{figure*}
\subsection{Timeliness-Fidelity Tradeoff with $\omega$-Wait Policy}\label{timeliness}
Figure~\ref{fig:big_res} illustrates the effect of the scheduling parameter $\omega_t \in (0, 120]$ under the $\omega$-wait policy on the performance of \gls{3d} scene representations, measured by the three image similarity metrics and $F_w$. In contrast to the $\omega$-Threshold strategy, the two $\omega$-wait curves exhibit no natural trade-off point. This is because the $\omega$-wait policy consistently relies on newly arrived frames, thereby avoiding the contamination from stale data. This can also be observed from the novel view synthesis result in Fig.~\ref{fig:results_compare}. As a result, PSNR and SSIM steadily increase and LPIPS decreases monotonically until convergence, with overall representation quality remaining higher across all $\omega_t$ values. 

Similar to the $\omega$-Threshold policy, in the high-traffic setting ($\lambda_g=\lambda_d=1/60$), the curves start from a lower baseline due to higher latency, rise more quickly as delivery rates improve, yet converge at $\omega_t = 62$ to a lower final quality; conversely, in the low-traffic setting ($\lambda_g=1/120, \lambda_d=1/30$), the curves start higher, increase more slowly, and ultimately achieve a higher steady-state quality at $\omega_t = 105$. A clear tradeoff emerges when the \gls{aaoi}-aware penalty $F_w$ is considered in Fig.~\ref{fig:big_res} (d); larger $\omega_t$ improves reliability by raising packet delivery probability, but simultaneously reduces freshness, leading to higher penalties in $F_w$. Thus, while $\omega$-wait avoids the degradation caused by stale data in representation quality metrics, it still reveals an inherent timeliness–reliability trade-off once freshness is explicitly evaluated.

Moreover, as illustrated in Fig.~\ref{fig:methods}, the timeliness–fidelity tradeoff consistently emerges across different \gls{3d} scene representation methods, highlighting the generality of this phenomenon beyond a single \gls{3d} scene representation method.

\begin{figure}
            \includegraphics[width=0.5\textwidth]{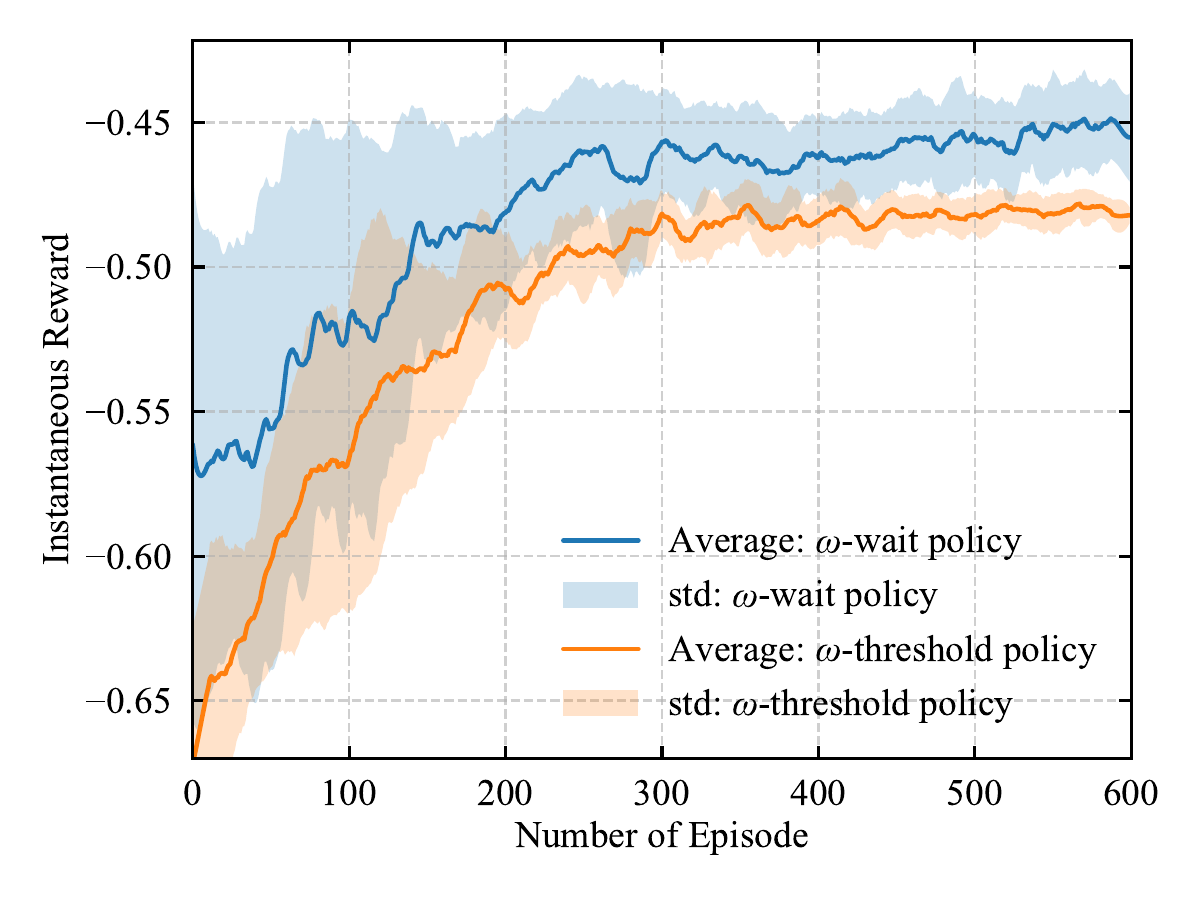}
           \caption{Training performance comparison of the two scheduling policies. The curves show the average instantaneous reward as a function of the training episode, while the shaded areas represent the standard deviation. Results are reported for the 
            $\omega$-wait and $\omega$-threshold policies.
           }
           \label{fig:reward}
\end{figure}

\subsection{\gls{3d} Scene Representations with Contextual-Bandit \gls{ppo}}\label{rl}

Figure~\ref{fig:reward} presents the training dynamics of our proposed \gls{rl} scheduler framework under the two scheduling policies in the \gls{spp} channel. Both policies exhibit steady performance improvement as training progresses, confirming the effectiveness of the reinforcement learning strategy in capturing the timeliness–fidelity tradeoff. Notably, the $\omega$-wait policy consistently outperforms the $\omega$-threshold policy, achieving a higher instantaneous reward at $r(\mathbf{s}_t, \mathbf{a}_t)=-0.46, \text{PSNR} =30.05, \text{SSIM} =0.793, \text{LPIPS} =0.248 $ after convergence. These results demonstrate that contextual-bandit \gls{ppo} can successfully learn an effective scheduling strategy to choose optimal $\omega_t$ for both policies, enabling real-time \gls{3d} scene representation to balance fidelity and timeliness more efficiently.


\section{Conclusion}
This paper examined the fundamental timeliness–fidelity tradeoff in real-time 3D scene representation over wireless networks. We analyzed two baseline scheduling strategies, $\omega$-threshold and $\omega$-wait, and validated the persistence of this tradeoff across multiple datasets and reconstruction methods. To overcome the limitations of these policies, we proposed a contextual-bandit \gls{ppo}-based scheduler that dynamically balances data freshness and reconstruction quality under an SPP channel. Experimental results demonstrated that the learned policy achieves superior fidelity while maintaining low  \gls{aoi}, highlighting its robustness and generality. These findings highlight the critical role of intelligent scheduling in real-time 3D scene representations and open opportunities for extending learning-based policies to heterogeneous sensing systems and more complex communication environments.

\bibliographystyle{IEEEtran}
\bibliography{bib}

\vspace{12pt}


\end{document}